\documentclass[10pt,conference]{IEEEtran}

\newif\iffinal
\finaltrue
\newcommand{\cmtid}{176}

\newif\ifarxiv
\arxivtrue

\newif\ifieee
\ieeefalse

\ifieee
\else
\usepackage[colorlinks=true,allcolors=black]{hyperref} %
\fi

\usepackage{icomma} %
\usepackage{cite}
\usepackage[acronym]{glossaries} %
\usepackage[capitalise]{cleveref} %
\usepackage{xspace}
\usepackage{booktabs}
\usepackage{arydshln}

\ifCLASSINFOpdf
   \usepackage[pdftex]{graphicx}
   \graphicspath{{figs/}}
   \DeclareGraphicsExtensions{.pdf,.jpeg,.png}
\else
   \usepackage[dvips]{graphicx}
   \graphicspath{{../figs/}}
   \DeclareGraphicsExtensions{.eps}
\fi

\usepackage{amsthm}

\usepackage{newtxtext,newtxmath} %

\usepackage{algorithmic}
\usepackage{array}

\ifCLASSOPTIONcompsoc
  \usepackage[caption=false,font=normalsize,labelfont=sf,textfont=sf]{subfig}
\else
  \usepackage[caption=false,font=footnotesize]{subfig}
\fi

\usepackage{multirow}
\usepackage{url}

\newcommand*{\RL}[2][]{\textcolor{Rhodamine}{[\textbf{\ifthenelse{\equal{#1}{}}{RL}{RL(#1)}}: #2]}}
\newcommand\RLI[1]{} %
\newcommand*{\DM}[2][]{\textcolor{blue}{[\textbf{\ifthenelse{\equal{#1}{}}{DM}{DM(#1)}}: #2]}}

\iffinal
\else
\usepackage[switch]{lineno}
\fi

\usepackage[dvipsnames,table]{xcolor}
\usepackage{balance}

\newboolean{showcomments}
\setboolean{showcomments}{true}     %

\newcommand*{\MyComment}[4][]{
  \ifthenelse{\boolean{showcomments}}{%
    \textcolor{#2}{[\textbf{\ifthenelse{\equal{#1}{}}{#3}{#3(#1)}}: #4]}%
  }{}%
}

\usepackage{cite}
\usepackage[normalem]{ulem}

\newacronymstyle{long-short-br}
{%
  \GlsUseAcrEntryDispStyle{long-short}%
}%
{%
  \GlsUseAcrStyleDefs{long-short}%
}
\setacronymstyle{long-short-br}

\newcounter{fncounter}
\newcommand\customfootnote[1]{\stepcounter{fncounter}\footnote{\hspace{0.2mm}#1}}

\usepackage{transparent}
\usepackage{tikz}
\ifarxiv
    \newcommand\copyrighttext{%
      \scriptsize Accepted at SIBGRAPI 2025. The final published version is available on IEEE Xplore (DOI: \href{https://doi.org/10.1109/SIBGRAPI67909.2025.11223367}{\textcolor{blue}{10.1109/SIBGRAPI67909.2025.11223367}}).}
    \newcommand\copyrightnotice{%
    \begin{tikzpicture}[remember picture,overlay]
    \node[anchor=south,yshift=30pt,xshift=0pt] at (current page.south) {\fbox{\transparent{0.85}\parbox{\dimexpr0.76\textwidth-\fboxsep-\fboxrule\relax}{\copyrighttext}}};
    \end{tikzpicture}%
    }
\else
\fi

\ifieee
\IEEEoverridecommandlockouts
\IEEEpubid{\makebox[\columnwidth]{979-8-3315-8951-6/25/\$31.00~\copyright2025 IEEE \hfill}
\hspace{\columnsep}\makebox[\columnwidth]{ }}
\else
\fi

\begin{document}
\title{\color{black} LPLC: A Dataset for License Plate\\Legibility Classification}

\iffinal
\author{\resizebox{0.99\linewidth}{!}{Lucas Wojcik\IEEEauthorrefmark{1},
Gabriel E. Lima\IEEEauthorrefmark{1},
Valfride Nascimento\IEEEauthorrefmark{1},
Eduil Nascimento Jr.\IEEEauthorrefmark{2},
Rayson Laroca\IEEEauthorrefmark{3}$^{,}$\IEEEauthorrefmark{1},
David Menotti\IEEEauthorrefmark{1}}\\

\IEEEauthorblockA{\IEEEauthorrefmark{1}Department of Informatics, Federal University of Paran\'a, Curitiba, Brazil}

\IEEEauthorblockA{\IEEEauthorrefmark{2}Department of Technological Development and Quality, Paran\'{a} Military Police, Curitiba, Brazil}

\IEEEauthorblockA{\IEEEauthorrefmark{3}Graduate Program in Informatics, Pontifical Catholic University of Paran\'a, Curitiba, Brazil}

\resizebox{0.99\linewidth}{!}{
\hspace{0.3mm}\IEEEauthorrefmark{1}{\hspace{-0.15mm}\tt\small \{lmlwojcik,gelima,vwnascimento,menotti\}@inf.ufpr.br} \quad \IEEEauthorrefmark{2}{\hspace{0.15mm}\tt\small eduiljunior@pm.pr.gov.br} \quad \IEEEauthorrefmark{3}{\hspace{0.15mm}\tt\small rayson@ppgia.pucpr.br}
}
}
\fi

\maketitle

\ifarxiv
    \copyrightnotice
\else
\fi

\newacronym{alpr}{ALPR}{Automatic License Plate Recognition}
\newacronym{gan}{GAN}{Generative Adversarial Network}
\newacronym{lp}{LP}{license plate}
\newacronym{dataset}{LPLC}{License Plate Legibility Classification}
\newacronym{ocr}{OCR}{Optical Character Recognition}
\newacronym{sr}{SR}{super-resolution}

\newcommand{\dataset}{\gls*{dataset}\xspace}
\newcommand{\parseq}{PARSeq-tiny\xspace}
\newcommand{\ocrchina}{GP\_LPR\xspace}

\newcommand{\StateName}{%
  \iffinal
    Paraná\xspace
  \fi
}

\newcommand{\urlDataset}{%
  \iffinal
    \url{https://github.com/lmlwojcik/lplc-dataset}\xspace
  \fi
}

\ifarxiv
  \vspace{-3mm}
\else
\fi

\begin{abstract}

\gls*{alpr} faces a major challenge when dealing with {\color{black}{illegible} \glspl*{lp}.
While reconstruction methods} such as \gls*{sr} have emerged, the core issue of {\color{black}recognizing these low-quality \glspl*{lp} remains unresolved.}
To optimize model performance and computational efficiency, image pre-processing should be {\color{black}applied selectively to cases that require enhanced legibility.
To support research in this area, we introduce a novel dataset comprising 10{,}210 images of vehicles with 12{,}687 annotated \glspl*{lp} for legibility classification (the \acrshort*{dataset} dataset).
The images span a wide range of vehicle types, lighting conditions, and camera/image quality levels.}
We adopt a fine-grained {\color{black}annotation strategy that includes vehicle- and LP-level occlusions, four legibility categories (perfect, good, poor, and illegible), and character labels for three categories (excluding illegible \glspl*{lp}).} 
As a benchmark, we propose a {\color{black}classification task using three image recognition networks to determine whether an \gls*{lp} image} is good enough, requires super-resolution, or is completely unrecoverable.
{\color{black}The overall F1 score, below 80\% for all three baseline models (ViT, ResNet, and YOLO), together with the analyses of \gls*{sr} and \gls*{lp} recognition methods, highlights the difficulty of the task and reinforces the need for further research.
}
The proposed dataset is publicly available at \urlDataset.

\end{abstract}

\IEEEpeerreviewmaketitle

\section{Introduction}
\glsresetall

\gls*{alpr} {\color{black}plays a vital role in road surveillance by using automated systems to detect vehicles and recognize them based on their \glspl*{lp}~\cite{laroca2021efficient,silva2022flexible}.
In the deep learning era, key research tasks in this domain include \gls*{lp} detection and \gls*{lp} recognition, often complemented by additional processes such as vehicle detection and \gls*{lp} rectification~\cite{ke2023ultra,ding2024endtoend}.}

{\color{black}State-of-the-art \gls*{alpr} methods achieve over 95\% accuracy in detection tasks and exceed 90\% recognition rates under ideal conditions~\cite{laroca2022cross,liu2024improving}.
However, their performance degrades significantly in challenging scenarios that are not well represented in most mainstream datasets.
Such scenarios include low-light environments, adverse weather conditions (e.g., rain), and low-resolution or low-quality images caused by poor equipment or transmission compression artifacts~\cite{wahyu2024fog, nascimento2025toward}.}

Furthermore, recent advances in \gls*{sr} technology have also opened up new paths to deal with faulty images~\cite{nascimento2024enhancing,pan2024lpsrgan}. 
These models were adapted to an \gls*{lp} reconstruction task and were shown to aid in recovering information from otherwise illegible images.
However, these models are often costly and not always needed, potentially even ruining otherwise suitable images for LP recognition (as we have observed in some of our experiments reported in this~work).

\begin{figure}[t]
    \centering
    \includegraphics[width=\linewidth]{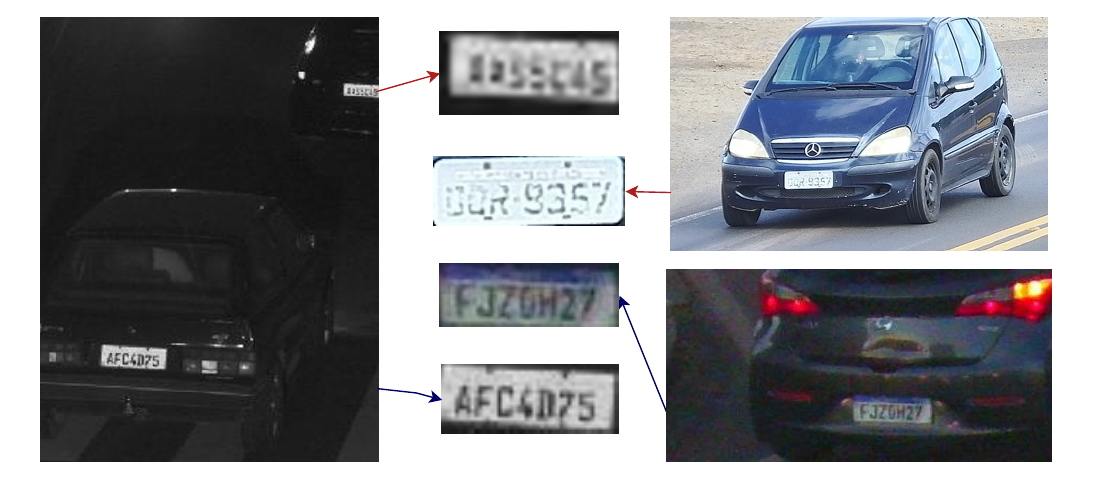}

    \vspace{-2mm}
    
    \caption{Examples illustrating the distinction between image quality and \gls*{lp} legibility. High-quality images may contain illegible \glspl*{lp}~(top right), while low-quality images can still include legible ones~(bottom right). A single image may also feature both legible and illegible~\glspl*{lp}~(left).}%
    \label{fig:quality-example}
\end{figure}

{\color{black}In this context, it is important to distinguish between image quality and character legibility. \gls*{lp} recognition performance depends not merely on image resolution or overall visual quality, but primarily on how legible the characters are.
\cref{fig:quality-example} illustrates this point: a visually high-quality image may contain illegible \glspl*{lp} due to factors such as camera distance, while a low-quality image may still feature clearly legible \glspl*{lp}.
In some cases, a single image can include both legible and illegible \glspl*{lp}.
We also observed that \gls*{ocr} models often produce incorrect predictions with high confidence~\cite{laroca2023leveraging}, particularly in cases where most characters are legible but one or two remain unclear~(an issue also depicted in \cref{fig:quality-example}).}

{\color{black}
With these challenges in mind, we present the \textit{\dataset} dataset\footnote{\color{black} The \dataset dataset is available at \urlDataset}, which contains over 10k radar images collected from various locations across the Brazilian state of \StateName and includes more than 12k annotated \glspl*{lp}.
The images exhibit a wide range of capture conditions, including varying lighting scenarios, vehicle types (motorcycles, cars, buses, and trucks), and image quality. 
A key contribution of this dataset is the assignment of each \gls*{lp} to one of four legibility levels: perfect, good, poor, or illegible.
These labels provide a qualitative and inherently subjective assessment of \gls*{lp} legibility.
For the first three categories, we also provide \gls*{ocr} annotations.
The dataset includes both Brazilian and Mercosur \glspl*{lp}\customfootnote{Following prior literature~\cite{silva2022flexible,laroca2023leveraging,nascimento2024enhancing}, we use the term ``Brazilian'' to refer to the \gls*{lp} layout used in Brazil prior to the adoption of the Mercosur~layout.} and can be used as a benchmark for various \gls*{alpr}-related~tasks.
}

In this work, we also use \dataset for a series of {\color{black}\gls*{lp} legibility} assessment tasks using established image classification models: ResNet~\cite{prajwal2023comparative}, ViT~\cite{dosovitskiy2021image} and YOLO-cls~\cite{ultralytics2025yolov11}.
{\color{black}Additionally, we explore the use of \gls*{sr} by evaluating three state-of-the-art models~\cite{nascimento2024enhancing,wang2021realesrgan,pan2024lpsrgan} on our novel dataset.
Building on these components, we design a recognition pipeline that incorporates a new decision step to determine whether an input image is sufficiently legible, requires further processing, or should be considered~unrecoverable.}

{\color{black}The remainder of this work is structured as follows. \cref{sec:related} provides an overview of state-of-the-art approaches in \gls*{lp} recognition, commonly used benchmark datasets, and key challenges. 
\cref{sec:dataset} introduces our proposed dataset, describing the images, \glspl*{lp}, and annotated attributes.
\cref{sec:experiments} describes our experimental setup for image legibility classification using established image processing models.
\cref{sec:results} presents and discusses the results.
Finally, \cref{sec:conclusion} concludes the paper.}

\section{Related Work}
\label{sec:related}

{\color{black}
State-of-the-art advancements in ALPR, as in other machine learning research domains, are quantitatively assessed using public benchmark datasets~\cite{laroca2022cross,ke2023ultra}.
An example is CLPD~\cite{clpd}, a dataset containing 1,200 images captured with various devices in mainland China.
While it primarily features daylight scenes and passenger cars, it offers considerable variability in backgrounds, road types, capture devices, and image quality.}

{\color{black}
For the Brazilian context, the RodoSol-ALPR dataset~\cite{laroca2022cross} comprises $20{,}000$ images captured at toll plazas, providing a realistic representation of actual operating conditions.
It includes a wide range of lighting scenarios, vehicle types, and weather conditions.
The authors~\cite{laroca2022cross} demonstrated that state-of-the-art methods perform well on older datasets featuring simpler scenarios, such as Caltech-Cars~\cite{caltech}, but struggle on RodoSol-ALPR, revealing limitations tied to outdated benchmarks.
While RodoSol-ALPR captures real situations, it is important to note that these toll plaza environments differ significantly from many surveillance camera settings, where camera quality and capture conditions are often much poorer and more~variable.}

{\color{black}
Recent advances in \gls*{lp} recognition have shifted focus from achieving high performance on simpler datasets to addressing more specific challenges in the field.
For example, Liu et al.~\cite{liu2024improving} proposed an attention-based decoder for \gls*{lp} recognition, following the growing trend of integrating attention mechanisms.
Their approach combines a CNN encoder for feature extraction with a contrastive learning strategy designed to differentiate both the position and class of each character.
The use of attention is motivated by the fixed character positions in \glspl*{lp} of a given format, enabling more effective layout-aware feature learning.
They reported a $96$\% recognition rate on~RodoSol-ALPR.}

{\color{black}
Similarly, Rao et al.~\cite{crnn} proposed an end-to-end pipeline that includes a segmentation step.
The method begins with a YOLOv5-based \gls*{lp} detector, followed by segmentation using the proposed AFF-Net model.
Next, a skew correction module is applied to the segmentation map using standard image processing techniques to extract the \gls*{lp} region.
The four corners of this region are then aligned with those of a canonical \gls*{lp} format.
Finally, \gls*{lp} recognition is performed using a CNN.
This correction step is effective in handling severely skewed \glspl*{lp} captured from oblique angles, such as those found in the CLPD dataset, where they reported a recognition rate of~\(94\%\).
}

{\color{black}
\Gls*{sr} has gained interest in \gls*{alpr} research due to its potential to recover otherwise illegible \gls*{lp} images.
For example, studies like~\cite{pan2024lpsrgan} adapt image-to-image translation methods, such as \gls*{gan}, to improve recognition performance on low-resolution \glspl*{lp}.
}

{\color{black}
In the same direction, Nascimento et al.~\cite{nascimento2024enhancing} proposed a novel focal loss tailored for character reconstruction, trained on a variant of the RodoSol-ALPR dataset called RodoSol-SR.
This dataset comprises pairs of high-resolution \gls*{lp} images and their synthetically degraded low-resolution counterparts.
These pairs were generated by the authors to train the model to reverse the degradation process through~reconstruction.
}

{\color{black}
Building on this context, we introduce the \dataset dataset, detailed in the following section.
Each \gls*{lp} is annotated with a qualitative legibility score using four distinct levels, which, to the best of our knowledge, is a novel contribution to the literature.
The dataset supports the development and evaluation of \gls*{lp} legibility classification methods, an important step in determining whether an image requires further processing such as super-resolution. It also serves as a more challenging benchmark for conventional LP detection and recognition~tasks.}

\section{The LPLC Dataset}
\label{sec:dataset}

\begin{figure*}
\centering

\resizebox{0.8\linewidth}{!}{
\includegraphics[height=0.2\textwidth]{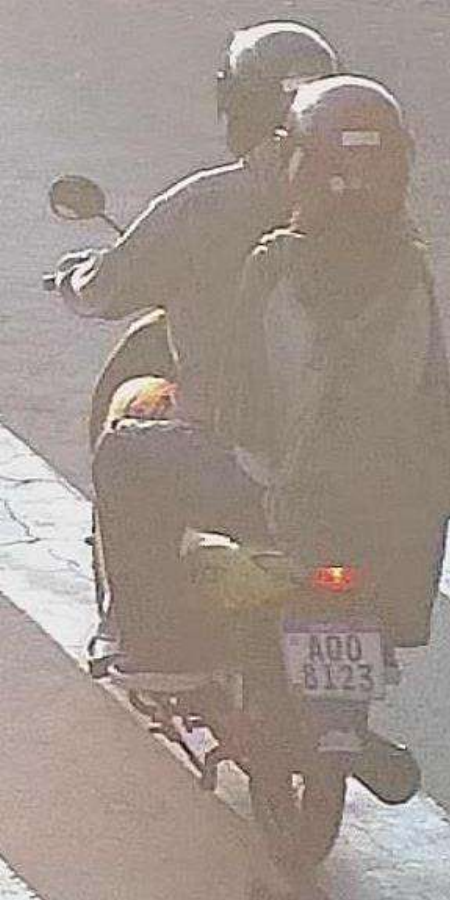}
\includegraphics[height=0.2\textwidth]{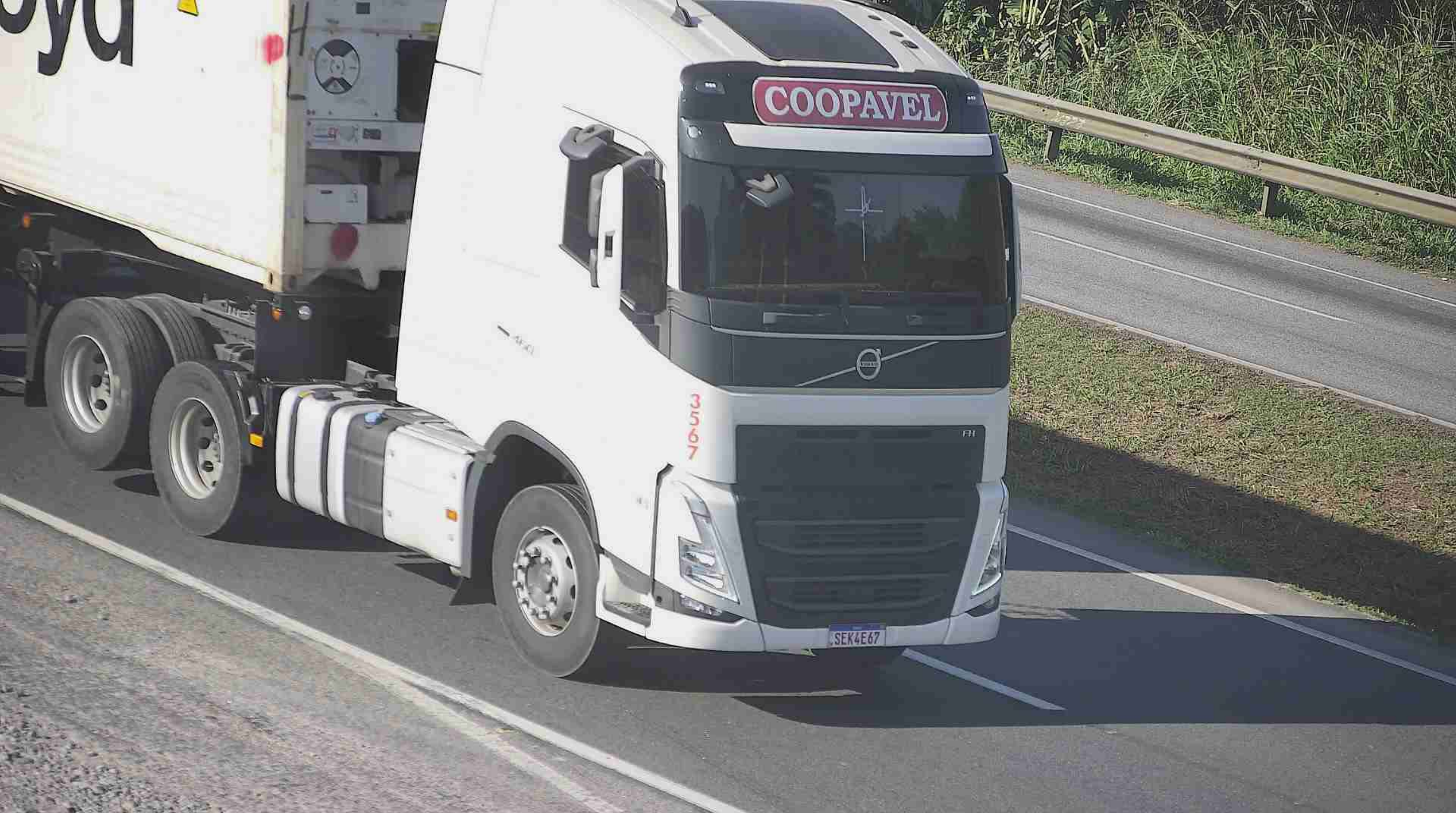}
\includegraphics[height=0.2\textwidth]{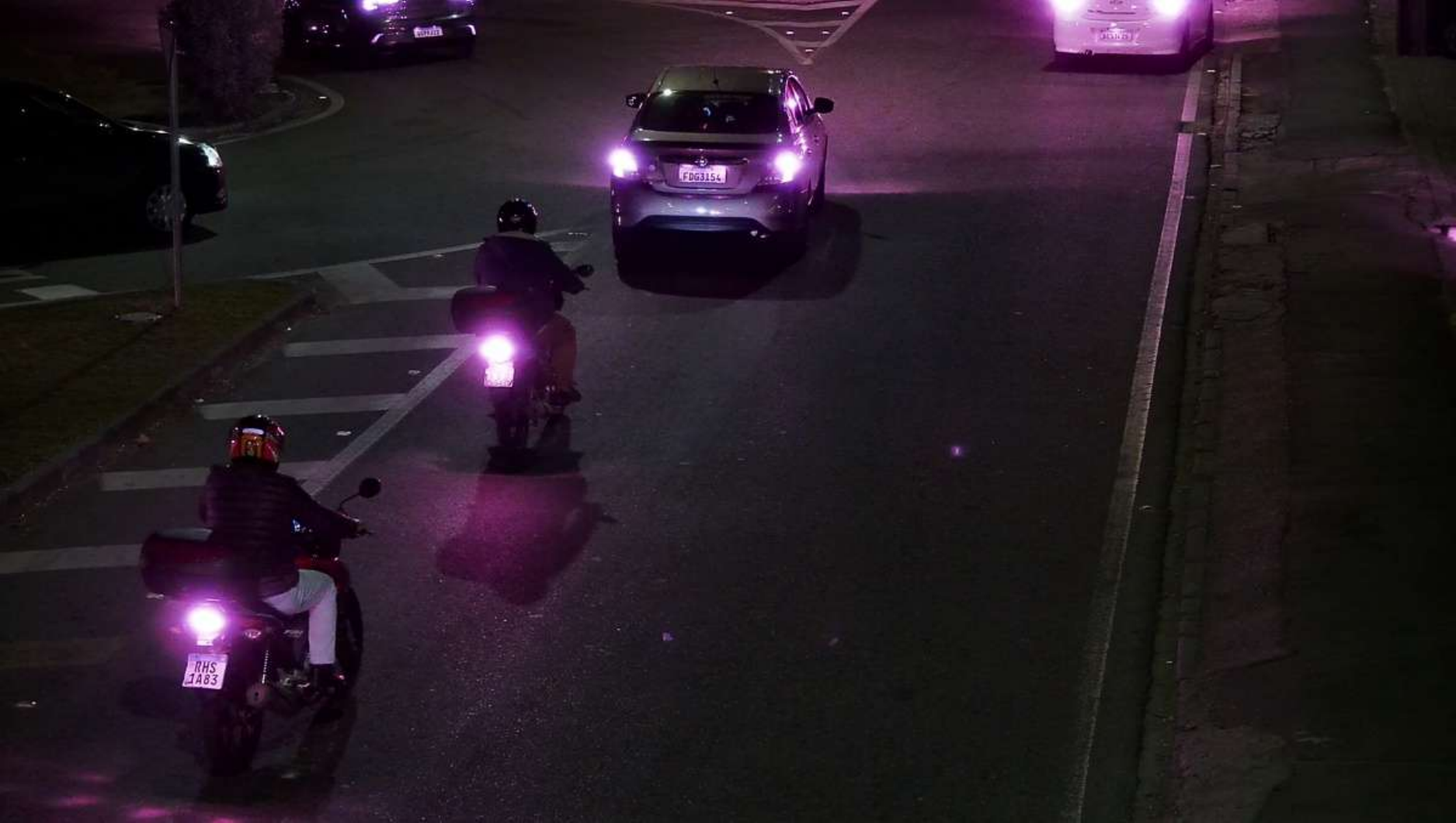}
}

\vspace{0.8mm}

\resizebox{0.8\linewidth}{!}{
\includegraphics[height=0.2\textwidth]{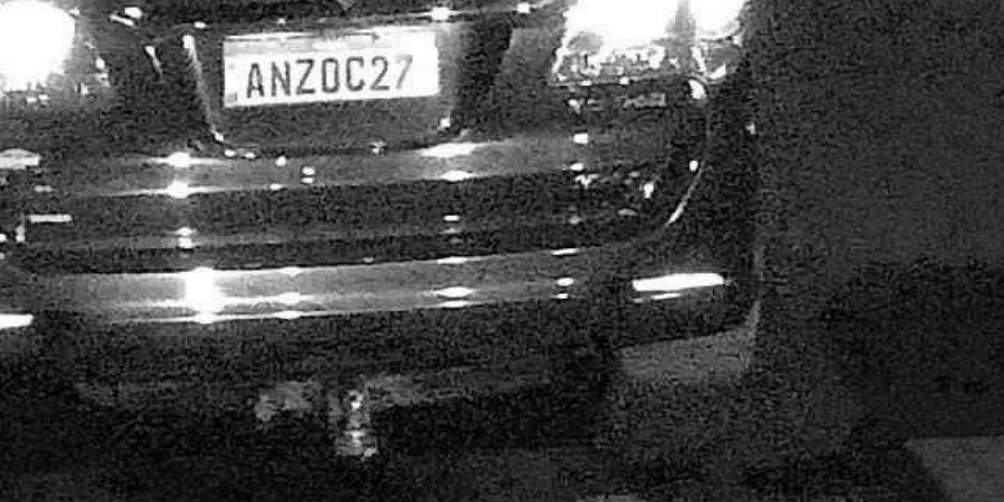}
\includegraphics[height=0.2\textwidth]{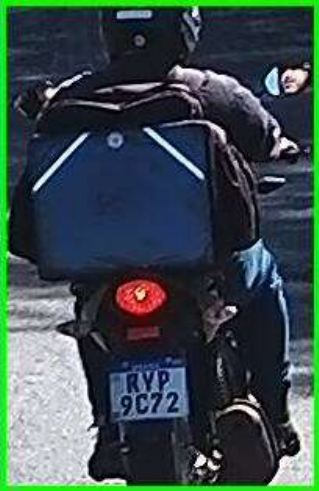}
\includegraphics[height=0.2\textwidth]{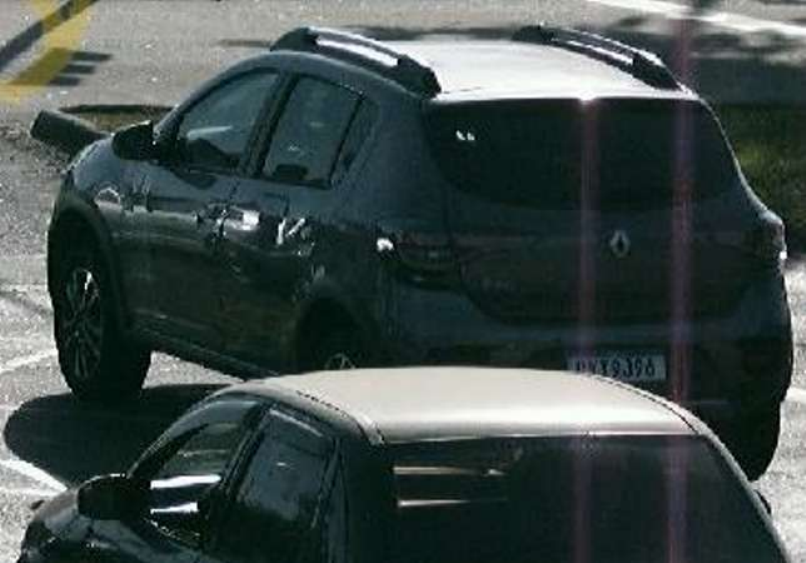}
}

\vspace{-2mm}

\caption{Samples from the \dataset dataset.}
\label{fig:lpqd_samples}
\end{figure*}

{\color{black}
We introduce the \textit{\acrfull*{dataset}} dataset, a publicly available resource primarily designed for the task of classifying the legibility of \gls*{lp} images, that is, whether an \gls*{lp} image is suitable for direct \gls*{ocr} or requires additional processing. 
Although its main purpose is legibility classification, the dataset is also suitable for other tasks such as \gls*{lp} detection and recognition, due to its fine-grained annotations that include details on \gls*{lp} text, legibility levels, and occlusion.}
{\color{black}All annotations were made manually by a single person (the first author) using the VIA annotation software~\cite{dutta2019vgg}, and validated through a semi-automatic process to ensure accuracy.
This revision process involved using an \gls*{ocr} model's outputs (\parseq~\cite{bautista2022scene}) and identifying legibility classification errors to flag potentially inconsistent annotations in both the legibility labels and \gls*{ocr} transcripts.}
\cref{fig:lpqd_samples} presents a few sample images from the~dataset.

{\color{black}
\dataset is composed of $10{,}210$ images captured by traffic radars across the Brazilian state of \StateName. 
All images were processed and redacted to remove metadata embedded by the cameras.}
These images were taken from hundreds of different cameras on various roads around the covered area. 
Each image is annotated with the time of capture (morning, afternoon, evening, or night) and may contain one or more \glspl*{lp}, totaling $12{,}687$ annotated \glspl*{lp}.
An \gls*{lp} is annotated if a significant part (roughly two-thirds) of the car is present in the image and if the \gls*{lp} {\color{black} is large enough to assess legibility.
Each \gls*{lp} is annotated with five attributes: (i)~the coordinates of the \gls*{lp}, (ii)~occlusion of the \gls*{lp}, (iii)~occlusion of the vehicle, (iv)~legibility level, and (v)~the \gls*{lp}~characters.}

{\color{black} The coordinates correspond to the four corners of the \gls*{lp} as it appears in the image, which may form an irregular quadrilateral}, {\color{black}starting at the top-left coordinate and moving clockwise towards the bottom-right.}
The \gls*{lp}-level occlusion is a binary {\color{black}attribute labeled ``occluded,'' which is set to true if one or more characters on the \gls*{lp} are not visible.
This often occurs when vehicles are positioned at the edge of the image but still meet our criteria for valid annotation, as seen in the bottom left vehicle in \cref{fig:lpqd_samples}.
Occlusion may also result from objects in the scene, such as other vehicles, tree branches, or lamp posts.
This is illustrated in the bottom right vehicle in \cref{fig:lpqd_samples}.
Vehicle-level occlusion is indicated by a binary attribute labeled ``valid.''
An \gls*{lp} is considered valid if it belongs to a vehicle with either most of its body (approximately 80\% visible or the full front or rear view, including both headlights or taillights) captured in the~image.}

{\color{black}
The annotation of \gls*{lp} characters is directly linked to their legibility level.
All \glspl*{lp} were manually labeled and then cross-validated using the \parseq \gls*{ocr} model~\cite{bautista2022scene}.
\glspl*{lp} classified as illegible ---~the lowest legibility level~--- are assigned an empty string, as their text cannot be recovered. 
For the remaining classes, the characters were annotated and validated through a semi-automatic process: the \gls*{ocr} model was used to predict each LP's text, and any discrepancies between the manual and predicted annotations were reviewed and corrected.
}

{\color{black}
The final validation step for potentially incorrect \gls*{ocr} annotations involved manually verifying whether the annotated \gls*{lp} text matched the vehicle shown in the image.
This was done by querying a paid API to retrieve vehicle information ---~such as make, model, color, and year~--- based on the annotated \gls*{lp} characters.
If the returned vehicle details did not match the one in the image, the annotation was reviewed and a new query was issued. This process was repeated until a correct match was found or up to four attempts had been made.
}

\begin{figure}[!htb]
\centering
\includegraphics[width=0.48\textwidth]{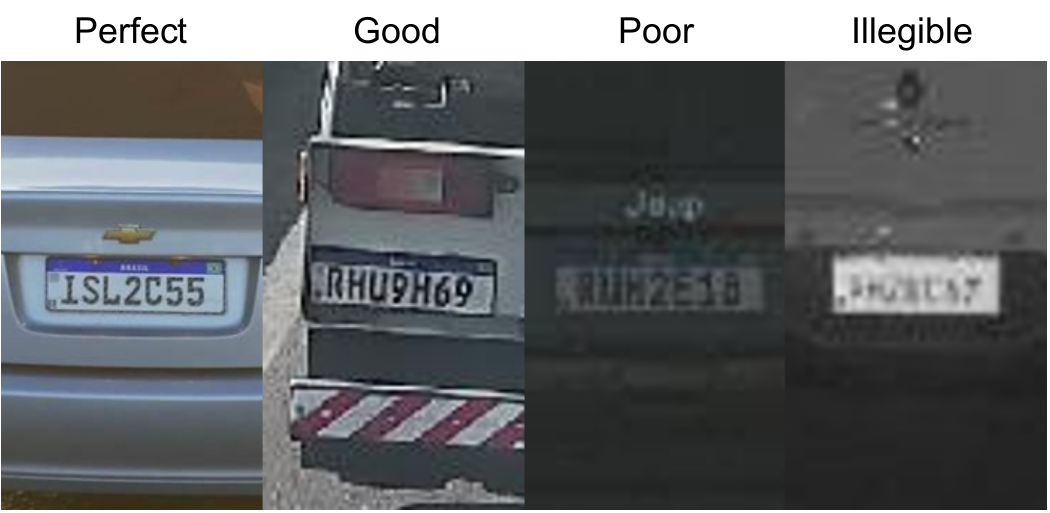}

\vspace{-2mm}

\caption{OCR legibility levels.}

\label{fig:ocr_levels}
\end{figure}

{\color{black}
The legibility level is annotated based on the visual quality of the \gls*{lp} text in the image.
We define four legibility levels, numbered from zero to three.
Class~0, labeled ``illegible,'' refers to \glspl*{lp} where the text is either completely unrecognizable or so degraded that it could not be validated using our method.
Class~1, ``poor,'' includes \glspl*{lp} with distorted text in which characters are not immediately recognizable.
Class~2, ``good,'' refers to \glspl*{lp} with legible text that may still exhibit some noise or visible distortion. 
Finally, Class~3, ``perfect,'' corresponds to \glspl*{lp} with clearly visible characters and no noticeable distortion. Examples of each legibility level are shown in~\cref{fig:ocr_levels}.
} %

{\color{black}
\cref{tab:plate_stats} shows the distribution of \glspl*{lp} and attributes in the dataset.
As indicated, the dataset is slightly imbalanced with respect to the legibility class.
An \gls*{ocr} annotation is considered true if the \gls*{lp}'s characters are labeled, even if some characters are missing due to occlusion.
As previously noted, all \glspl*{lp} have annotated characters, except those categorized as~illegible.
}

\begin{table}[!htb]
    \renewcommand{\arraystretch}{1.05}
    \caption{LP Statistics.}%
    \label{tab:plate_stats}

    \vspace{-2mm}
    
    \centering
    \begin{tabular}{lclcc}
        \toprule
        \multicolumn{2}{c}{LPs by Legibility} & \multicolumn{3}{c}{Other Attributes} \\
        \cmidrule(lr){1-2} \cmidrule(lr){3-5}
        Class & Number & Class & True & False \\
        \midrule
        Perfect    & $5{,}617$ & Occluded   & $12{,}586$ & $\phantom{0}\phantom{,}101$ \\
        Good       & $3{,}641$ & Valid      & $12{,}359$ & $\phantom{0}\phantom{,}328$ \\
        Poor       & $1{,}825$ & OCR        & $11{,}083$ & $1{,}604$ \\
        Illegible  & $1{,}604$ & $\rightarrow$ Total LPs  & \multicolumn{2}{c}{$12{,}687$} \\
        \bottomrule

    \end{tabular}
\end{table}

\cref{tab:image_stats} presents the distribution of images in the dataset.
{\color{black}
While nighttime images constitute a minority of our dataset ($4{,}665$ evening and nighttime images versus $6{,}386$ morning and afternoon images), this represents a significant contribution compared to other public datasets, which typically contain minimal nighttime imagery~\cite{onim2023unleashing}.}
{\color{black}
Most images contain at least one vehicle with a positive label (an LP that is legible, non-occluded, or associated with a valid vehicle), and $9{,}454$ images include at least one \gls*{lp} that is legible, valid, and not~occluded.}

\begin{table}[!htb]
\renewcommand{\arraystretch}{1.05}
\caption{Image Statistics.}
\label{tab:image_stats}

\vspace{-2mm}

\centering
\begin{tabular}{lclc}
\toprule
\multicolumn{2}{c}{Images by Time of Day} & \multicolumn{2}{c}{Images by Attributes} \\
\cmidrule(r){1-2} \cmidrule(l){3-4}
Class & Number & Has at Least One & Number \\
\midrule
Morning   & $3{,}830$  & Legible LPs      & $\phantom{0}9{,}684$ \\
Afternoon & $2{,}556$  & Non Occluded LPs & $10{,}195$ \\
Evening   & $2{,}585$  & Valid Vehicles   & $10{,}030$ \\
Night     & $1{,}239$  & $\rightarrow$ Total Images     & $10{,}200$ \\
\bottomrule
\end{tabular}
\end{table}

\section{Experimental Setup}
\label{sec:experiments}

{\color{black}
Our experiments focus on the legibility attribute at the \gls*{lp} level.
We crop all $12{,}687$ \glspl*{lp} from the source images using their annotated corner coordinates.
To evaluate legibility classification, we consider three models: ResNet~\cite{he2016deep}, ViT~\cite{dosovitskiy2021image} and YOLO-cls~\cite{ultralytics2025yolov11}.
The goal is to assess whether an \gls*{lp} image is suitable for \gls*{ocr} processing, enabling more efficient downstream \gls*{lp} recognition by filtering out low-quality~samples.}

\begin{figure}
\centering
\includegraphics[width=0.48\textwidth]{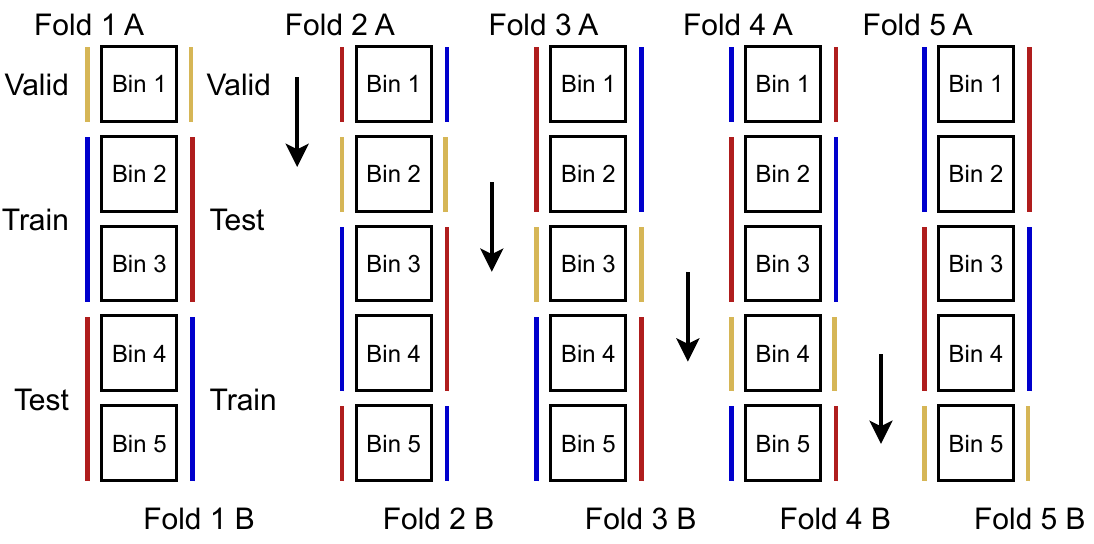}
\vspace{-2mm}
\caption{Cross-fold Splits Illustration.}
\label{fig:fold_splits}
\end{figure}

We employ a $5$-fold cross-validation protocol with {\color{black} $40\%$, $20\%$, and $40\%$ splits for training, validation, and testing, respectively.}
Each fold generates two different experiment iterations, where the training and test partitions are flipped, as illustrated in \cref{fig:fold_splits}.
{\color{black} Consequently}, each experiment is carried out \(10\)~times, once per unique fold configuration, and the final results are {\color{black}reported} as the average over these \(10\)~runs.

There are {\color{black}several} reasons behind our choice of protocol.
First, the $n$-fold cross-validation protocol allows us to use the entire dataset for testing, {\color{black}providing} performance results for every image.
Second, it mitigates bias {\color{black}associated with specific training and test partitions.
Averaging results across all folds ensures that the evaluation is not influenced by a particularly easy or difficult split.
Lastly, alternating the roles of training and test partitions between rounds allows for more accurate comparisons across folds.
The fold splits are made available alongside the~dataset.}

{\color{black}
We train each model under three scenarios, all framed as straightforward image classification tasks using the cropped \glspl*{lp}.
The first scenario, called ``Baseline,'' involves predicting one of the four standard legibility classes: perfect, good, poor, or illegible.
This setup aims to evaluate how effectively current models can assess the legibility of a given \gls*{lp}~image.
}

{\color{black}
In the second scenario, termed ``Legibility Recognition'', we merge the \textit{perfect} and \textit{good} \glspl*{lp} into a single class ``legible'' and train the network to perform a binary classification between legible and poor \glspl*{lp}.
\textit{Illegible} \glspl*{lp} are excluded from this experiment. 
The objective is to determine whether an image is suitable for \gls*{ocr} or requires further processing ---~hence the binary output.
The third scenario, ``Full Recognition'', builds on the second by reintroducing \textit{illegible} \glspl*{lp} as a third class, framing the task as ternary classification.
In addition to identifying images that may or may not require enhancement for \gls*{ocr}, this scenario also aims to detect unrecoverable images that should be discarded. %
}

{\color{black}
We also assess the impact of \gls*{sr} on \gls*{lp} recognition performance.
For \gls*{sr}, we adopt the pipelines proposed in \cite{nascimento2024enhancing}, where images are first reconstructed using their \gls*{sr} model and then processed with a version of the \ocrchina \gls*{ocr} model~\cite{liu2024irregular} trained on the RodoSol-ALPR dataset~\cite{laroca2022cross}.
Both \gls*{sr} and \gls*{ocr} are applied to all images, and the OCR results from the reconstructed images are compared against those from the originals.
In this experiment, in addition to the LCOFL-GAN proposed in~\cite{nascimento2024enhancing}, we also evaluate Real-ESRGAN~\cite{wang2021realesrgan} and LPSRGAN~\cite{pan2024lpsrgan}, providing a broader overview of current \gls*{sr} state of the~art.}

{\color{black}
Finally, we report OCR results by legibility class.
In addition to \ocrchina, we also evaluate the \parseq model~\cite{bautista2022scene}, trained on an unpublished dataset. %
We report both character-level and \gls*{lp}-level recognition rate, presenting the results broken down by legibility~class.
}

\subsection{Fine-Tuning Parameters}

{\color{black}
As mentioned before, we train ResNet, YOLO-cls, and ViT in these scenarios. 
Specifically, we employ ResNet-50 and ViT Base-16 from PyTorch's torchvision library, and YOLO11m-cls from the ultralytics implementation.
These models were selected based on their proven effectiveness and widespread adoption in state-of-the-art research across diverse domains~\cite{yolo_app,lima2024toward,laroca2025improving}.}
We initialize all models with the weights pre-trained on ImageNet, and swap the last layer for a simple linear layer with \(N\) outputs, where \(N\) is the number of classes in a given training scenario~($4$, $2$, and $3$ classes for {\color{black}scenarios \(1\), \(2\), and \(3\),~respectively).}

{\color{black}
All experiments are conducted on an NVIDIA GeForce 3090 RTX GPU.
Each model is trained across ten folds for a maximum of $200$ epochs with a batch size of $16$.}
We employ an early stopping strategy based on the validation set accuracy, with a patience of $20$ epochs.
For ViT and ResNet, we use the Adam optimizer at a learning rate of $10^{-5}$.
YOLO-cls is trained with an SGD optimizer, a starting learning rate of $10^{-2}$, multiplied by $10^{-2}$ following the cosine weight decay.
For all models, we fine-tune every layer instead of only the final classifier~layer.

\section{Results}
\label{sec:results}

\begin{figure}[!htb]
    \centering
    
    \hspace{-1.5mm}\includegraphics[width=0.625\linewidth]{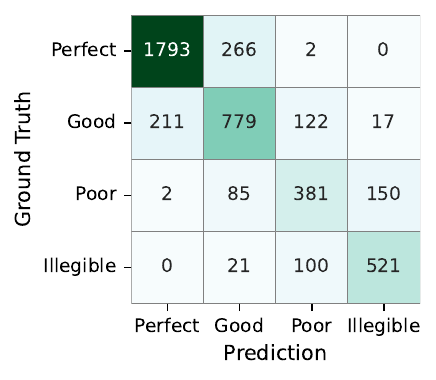}

    \vspace{-4mm}
    
    \caption{Confusion matrix for one run of YOLO-cls.}
    \label{fig:confusion}
\end{figure}

\cref{tab:scen_base} presents our results on the {\color{black}\textit{Baseline}} scenario, which consists of predicting the {\color{black}legibility} label for a given LP image.
We report the average test micro-F1 score across the $10$ folds for each class, {\color{black}along with the overall F1 score.
Despite the task's straightforward nature and limited number of classes, the results reveal it to be quite challenging.
Among the models evaluated, YOLO-cls achieves the best performance.
\cref{fig:confusion} shows the confusion matrix of YOLO-cls for one of the folds, revealing substantial overlap between adjacent classes and blurred decision boundaries ---~a pattern that corresponds to the ambiguous cases where the model made the most~errors.}

\begin{table}[!htb]
    \centering
    \renewcommand{\arraystretch}{1.05}
    \caption{Results on the Baseline scenario (F1-score).}
    
    \vspace{-2mm}
    
    \begin{tabular}{lccccc}
        \toprule
        \multirow{2}[2]{*}{Model} & \multicolumn{4}{c}{Class} & \multirow{2}[2]{*}{Overall} \\
        \cmidrule(lr){2-5}
        & Perfect & Good & Poor & Illegible & \\
        \midrule
        ResNet-50    & $84.54$\% & $67.98$\% & $56.70$\% & $72.97$\% & $74.51$\% \\
        ViT b-16     & $85.74$\% & $68.00$\% & $58.80$\% & $73.67$\% & $75.48$\% \\
        YOLO11m-cls  & $88.37$\% & $65.83$\% & $59.42$\% & $74.47$\% & $76.79$\% \\
        \bottomrule
    \end{tabular}
    \label{tab:scen_base}
\end{table}

\cref{tab:scen_12} presents the classification results for the \textit{Legibility Recognition} and \textit{Full Recognition} scenarios.
The first scenario evaluates whether an \gls*{lp} image is {\color{black}suitable for \gls*{ocr}}, while the second {\color{black}introduces} a third class to identify \glspl*{lp} with unrecoverable {\color{black}text}.
We report the test micro-F1 score averaged across the \(10\) folds.
{\color{black}Due to the reduced number of classes, this becomes a simpler task in which the model distinguishes between legible and non-legible~(or unrecoverable) images.
As expected, the classification metrics are correspondingly higher.
Nevertheless, there remains considerable room for improvement, as the models consistently struggle to differentiate between high- and low-quality textures across all~scenarios.}

\begin{table}[!htb]
    \centering
    \renewcommand{\arraystretch}{1.05}
    \caption{Results on the Legibility scenario (F1-score).}

    \vspace{-2mm}
    
    \begin{tabular}{lcc}
        \toprule
        \multirow{2}{*}{Model}
            & Legibility Recognition & Full Recognition \\
            & (Legible vs.\ Poor) & (Legible, Poor, Illegible) \\
        \midrule
        ResNet-50    & $92.56$\% & $87.23$\% \\
        ViT b-16     & $93.16$\% & $87.78$\% \\
        YOLO11m-cls  & $92.71$\% & $86.25$\% \\
        \bottomrule
    \end{tabular}
    \label{tab:scen_12}
\end{table}

{\color{black}\cref{tab:sr_eval} presents the performance of the pre-trained \gls*{sr} models when integrated with \gls*{lp} recognition.
All cropped \glspl*{lp} from the \dataset dataset were first enhanced using the respective \gls*{sr} model (LCOFL-GAN~\cite{nascimento2024enhancing}, RealESRGAN~\cite{wang2021realesrgan}, and LPSRGAN~\cite{pan2024lpsrgan}) and then processed with the \ocrchina \gls*{ocr} model~\cite{liu2024irregular}, trained on the RodoSol-ALPR dataset~\cite{laroca2022cross} (as in~\cite{nascimento2024enhancing}).
This setup evaluates whether \gls*{sr} improves \gls*{lp} recognition accuracy.
The results suggest otherwise: in the best case (Real-ESRGAN), recognition improved for only~\(647\) out of the~\(11{,}027\) \glspl*{lp}~(\(\le6\%\)). These findings indicate that current \gls*{sr} networks struggle to generalize across~datasets.}

\begin{table}[!htb]
    \centering
    \renewcommand{\arraystretch}{1.05}
    \caption{SR network evaluation (character accuracy).}

    \vspace{-2mm}

    \resizebox{0.99\linewidth}{!}{
    \begin{tabular}{lccccc}
        \toprule
        \multirow{2}[2]{*}{GAN Model} & \multirow{2}[2]{*}{\begin{tabular}{@{}c@{}}OCR Results \\ With SR\end{tabular}} & \multicolumn{3}{c}{Class} & \multirow{2}[2]{*}{\phantom{0}Total} \\
        \cmidrule(lr){3-5}
        & & Perfect & Good & Poor & \\
        \midrule
        \multirow{3}{*}{LCOFL-GAN~\cite{nascimento2024enhancing}} & Better & $\phantom{000}98$ & $\phantom{0,}108$ & $\phantom{0,}108$ & $\phantom{00,}314$  \\
        & Equal & $\phantom{00}494$ & $\phantom{0,}308$ & $\phantom{0,}170$ & $\phantom{00,}972$ \\
        & Worse & $5{,}012$ & $3{,}206$ & $1{,}523$ & $\phantom{0}9{,}741$ \\
        \midrule
        {\color{black}\multirow{3}{*}{Real-ESRGAN~\cite{wang2021realesrgan}}} & Better & $\phantom{00}211$ & $\phantom{0,}276$ & $\phantom{0,}160$ & $\phantom{00,}647$ \\
        & Equal & $5{,}180$ & $2{,}683$ & $\phantom{0,}572$ & $\phantom{0}8{,}435$ \\
        & Worse & $\phantom{0,}213$ & $\phantom{0,}663$ & $1{,}069$ & $\phantom{0}1{,}945$ \\
        \midrule
        {\color{black}\multirow{3}{*}{LPSRGAN~\cite{pan2024lpsrgan}}} & Better & $\phantom{0,}108$ & $\phantom{000}98$ & $\phantom{0,0}45$ & $\phantom{00,}251$  \\
        & Equal & $\phantom{,0}264$ & $\phantom{00}209$ & $\phantom{0,}115$ & $\phantom{00,}588$ \\
        & Worse & $5{,}232$ & $3{,}315$ & $1{,}641$ & $10{,}188$ \\
        \bottomrule
    \end{tabular}
    }
    \label{tab:sr_eval}
\end{table}

\begin{figure}[!htb]
    \centering
    \includegraphics[width=0.8\linewidth]{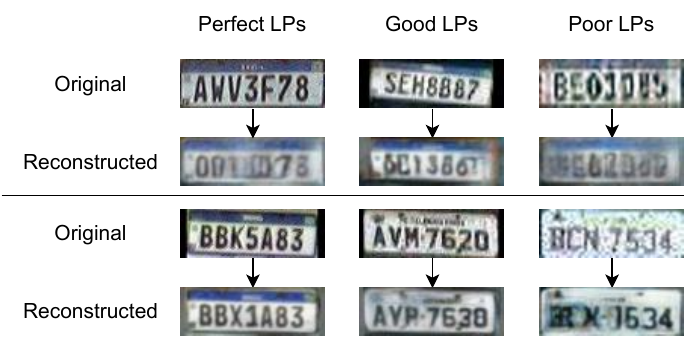}
    \caption{Original and reconstructed images using super-resolution.}
    \label{fig:lp_sr}
\end{figure}

{\color{black}This demonstrates that applying super-resolution does not necessarily enhance \gls*{lp} recognition performance and can, in many cases, be detrimental.
\cref{fig:lp_sr} shows six reconstructed images (using LCOFL-GAN) alongside their original versions, organized by legibility level.
As can be seen, the employed \gls*{sr} model not only reduces legibility but also introduces hallucinated characters that were not present in the original~images.}

Finally, \cref{tab:ocr_results} presents the character-wise accuracy~(Char Acc) and {\color{black}whole-\gls*{lp} accuracy~(\gls*{lp} Acc) for our dataset, broken down by legibility class.
As described in \cref{sec:experiments}}, we use \ocrchina~\cite{liu2024irregular} and \parseq~\cite{bautista2022scene}, the latter of which was trained on a different, unpublished dataset.
{\color{black}The results clearly indicate that \gls*{ocr} performance is strongly correlated with the legibility of the \gls*{lp}: lower legibility levels lead to significantly reduced recognition~accuracy.}

\begin{table}[!htb]
    \centering
    \renewcommand{\arraystretch}{1.05}
    \caption{\gls*{ocr} evaluation on the \dataset dataset.}
    \label{tab:ocr_results}
    
    \vspace{-2mm}
    
    \begin{tabular}{lcccc}
        \toprule
        \multirow{2}[2]{*}{Class} & \multicolumn{2}{c}{GP\_LPR~\cite{liu2024irregular}} & \multicolumn{2}{c}{\parseq~\cite{bautista2022scene}} \\
        \cmidrule(r){2-3} \cmidrule(l){4-5}
        & Char Acc & \gls*{lp} Acc & Char Acc & \gls*{lp} Acc \\
        \midrule
        Perfect & $90.57$\% & $80.74$\% & $99.52$\% & $98.07$\% \\
        Good    & $85.02$\% & $62.37$\% & $98.40$\% & $92.32$\% \\
        Poor    & $74.91$\% & $30.30$\% & $93.78$\% & $72.34$\% \\
        \midrule
        Overall & $86.30$\% & $66.84$\% & $98.08$\% & $91.37$\% \\
        \bottomrule
    \end{tabular}
\end{table}

\vspace{1mm}
\section{Conclusions}
\label{sec:conclusion}

{\color{black}
In this work, we introduced a novel \gls*{alpr}-related dataset consisting of $10{,}200$ images captured by street radars, encompassing a diverse range of vehicles and \glspl*{lp}.
The dataset includes fine-grained annotations such as legibility level, \gls*{lp} text, and both vehicle- and \gls*{lp}-level occlusion.
We adapted three image classification models for the task of legibility classification and evaluated them in a \(5\)-fold cross-validation setup, reporting average results across ten runs.
}

{\color{black} Our findings indicate that although the evaluated classification networks achieve promising results, they are not yet reliable for deployment in real-world applications, particularly due to their difficulty in distinguishing subtle legibility differences. Furthermore, the super-resolution model evaluated in this study exhibits poor generalization in cross-dataset settings, frequently degrading \gls*{lp} legibility or introducing hallucinated characters, ultimately harming \gls*{lp} recognition performance.

As future work, we suggest the development of \gls*{sr} models tailored for cross-domain scenarios, as well as the exploration of alternative image enhancement techniques aimed at improving \gls*{lp} legibility without compromising the integrity of the visual content.
In addition, we plan to position this dataset within the state of the art by conducting a more comprehensive comparative evaluation against current \gls*{alpr} methods, thereby demonstrating its broad~applicability.
}

\iffinal
\balance
\else
\fi

\section*{Acknowledgments}

\iffinal
This study was financed in part by the \textit{Coordenação de Aperfeiçoamento de Pessoal de Nível Superior - Brasil~(CAPES)}, through the \textit{Programa de Excelência Acadêmica (PROEX)} - Finance Code 001, and in part by the \textit{Conselho Nacional de Desenvolvimento Científico e Tecnológico~(CNPq)}~(\#~315409/2023-1). The authors gratefully thank these organizations for the support.
\fi

\bibliographystyle{IEEEtran}
\bibliography{example}

\end{document}